\def\year{2022}\relax
\documentclass[letterpaper]{article} 
\usepackage{aaai22}  
\usepackage{times}  
\usepackage{helvet}  
\usepackage{courier}  
\usepackage[hyphens]{url}  
\usepackage{graphicx} 
\usepackage{natbib}  
\usepackage{caption} 
\DeclareCaptionStyle{ruled}{labelfont=normalfont,labelsep=colon,strut=off} 
\usepackage{algorithm}
\usepackage{algorithmic}

\usepackage{newfloat}
\usepackage{listings}
\floatstyle{ruled}
\newfloat{listing}{tb}{lst}{}
\floatname{listing}{Listing}
\usepackage{amsfonts}
\usepackage{amsmath}
\usepackage{amssymb}
\usepackage{booktabs}
\usepackage{xcolor}

\title{Representing Prior Knowledge Using Randomly, Weighted Feature Networks for Visual Relationship Detection}
\author {
    Jinyung Hong\textsuperscript{\rm 1},
    Theodore P. Pavlic\textsuperscript{\rm 1,2,3,4}
}
\affiliations {
    \textsuperscript{\rm 1} School of Computing and Augmented Intelligence, Arizona State University, Tempe, AZ 85281, USA\\
    \textsuperscript{\rm 2} School of Sustainability, Arizona State University, Tempe, AZ 85281, USA\\
    \textsuperscript{\rm 3} School of Complex Adaptive Systems, Arizona State University, Tempe, AZ 85281, USA\\
    \textsuperscript{\rm 4} School of Life Sciences, Arizona State University, Tempe, AZ 85281, USA\\
    jhong53@asu.edu,
    tpavlic@asu.edu
}

\begin{document}

\maketitle

\begin{abstract}
The single-hidden-layer Randomly Weighted Feature Network~(RWFN) introduced by \citet{hong2021insect} was developed as an alternative to neural tensor network approaches for relational learning tasks. Its relatively small footprint combined with the use of two randomized input projections~-- an insect-brain--inspired input representation and random Fourier features~-- allow it to achieve rich expressiveness for relational learning with relatively low training cost. In particular, when \citeauthor{hong2021insect} compared RWFN to Logic Tensor Networks~(LTNs) for Semantic Image Interpretation~(SII) tasks to extract structured semantic descriptions from images, they showed that the RWFN integration of the two hidden, randomized representations better captures relationships among inputs with a faster training process even though it uses far fewer learnable parameters. In this paper, we use RWFNs to perform Visual Relationship Detection~(VRD) tasks, which are more challenging SII tasks. A zero-shot learning approach is used with RWFN that can exploit similarities with other seen relationships and background knowledge~-- expressed with logical constraints between subjects, relations, and objects~-- to achieve the ability to predict triples that do not appear in the training set. The experiments on the Visual Relationship Dataset to compare the performance between RWFNs and LTNs, one of the leading Statistical Relational Learning frameworks, show that RWFNs outperform LTNs for the predicate-detection task while using fewer number of adaptable parameters~($1:56$ ratio). Furthermore, background knowledge represented by RWFNs can be used to alleviate the incompleteness of training sets even though the space complexity of RWFNs is much smaller than LTNs~($1:27$ ratio).
\end{abstract}

\section{Introduction}
Semantic Image Interpretation~(SII)~\citep{neumann2008scene} is a task of automatically extracting high-level information about the content of a visual scene. This information consists of the objects in the scene, their attributes, and the relations among them. Formally, the result of applying SII on an image is called a \emph{scene graph}~\citep{krishna2017visual}: the labeled nodes are regarded as objects in the scene and their attributes, the labeled edges indicate the relations between the corresponding nodes. The resulting scene graphs can be applied to many application domains, such as visual question answering, image captioning, image querying, and robot interaction.

Visual Relationship Detection~(VRD) tasks are a special case of scene-graph construction, and a visual relationship can be readily represented as a triple of the form $\langle s, p, o \rangle$ where $s$ and $o$ indicate the subject and the object, respectively, that are the semantic classes or labels of two bounding boxes in the image. The predicate $p$ is the label representing the relationship between the two bounding boxes. Therefore, constructing a scene graph in a visual scene is the task of correctly labeling subjects and objects nodes in the graph and labeling edges connected from subject to object.

In VRD tasks, visual relationships are mainly detected in a supervised fashion, but there are some challenges. For one, due to the enormous human effort of the annotators, a comprehensive and detailed annotation is not achievable. Furthermore, even when a training set covers a variety of different relationships, a new type of relationship may appear only in the test set. We use \emph{zero-shot learning}~\citep{lampert2013attribute} to address these VRD issues. The zero-shot approach can be achieved by exploiting the similarity with the triples in the training set or a high-level description of the relationship. This is closer to human learning for supervised learning. Indeed, humans can both generalize from seen or similar examples and use their background knowledge to identify never seen relationships~\citep{lampert2013attribute}.

A zero-shot training approach using Logic Tensor Networks~ (LTN)~\citep{donadello2019compensating} has been proposed as one of the state-of-the-art methods for detecting invisible visual relationships. The LTN~\citep{serafini2016learning} is a Statistical Relational Learning framework that learns from relational data in the presence of additional logical constraints. So, \citet{donadello2019compensating} leveraged LTNs to exploit the similarities with already seen triples. They showed that the results on the Visual Relationship Dataset jointly using logical knowledge and data outperform the state-of-the-art approaches based on data or linguistic knowledge and that logical knowledge can compensate for the incompleteness of the datasets due to the high annotation effort.

Recently, a more economical approach for representing background knowledge has been proposed that is inspired by the neural architecture of the insect brain. Insect neuroscience has shown that insects express sophisticated and complex behaviors although they possess central nervous systems far smaller than the human brain~\citep{avargues2011visual}. Specifically, it has been shown that the honey bee brain includes high levels of cognitive sophistication to learn relational concepts such as ``same,'' ``different,'' ``larger than,'' ``better than,'' among others~\citep{avargues2013conceptual}. Based on those observations, \citet{hong2021insect} proposed a novel insect-brain--inspired neural network, Randomly Weighted Feature Network~(RWFN), for relational embedding that incorporates randomly drawn, untrained weights in its encoder with a trained linear model as a decoder. Their approach mimics the randomized, mostly feedforward architecture of the insect brain that projects olfactory features from the antennal lobe~(AL) randomly to the mushroom body~(MB), the main center of higher-order learning, that are then decoded by downstream neuropils that translate information processing into action. In addition to the insect-inspired architecture, RWFN also leverages random Fourier feature~\citep{rahimi2007random}, a kernel approximation method that overcomes the issues of conventional kernel machines or kernel methods~\citep{smola1998learning} to concisely and efficiently compute linear interactions among inputs. When RWFNs were applied to one of the SII tasks defined by~\citet{donadello2017logic} to demonstrate the performance of LTNs, the RWFNs were able to effectively learn the \emph{part-of} relation among inputs better than the neural tensor network in LTNs. In addition, \citet{hong2021insect} used an ablation study to show that the two randomized representations in RWFNs can be compensated for each other to achieve balanced performance among different kinds of dataset.

In this paper, we extend RWFNs to address zero-shot learning for the detection of unseen visual relationships. We perform experiments on the Visual Relationship Dataset~(VRD)~\citep{lu2016visual}, a complex dataset containing 100 unary relationships and 70 binary relationships, and compare the performance between RWFNs and LTNs. 

\section{Related Work}
\subsection{Methods for Visual Relationship Detection}
In order to extract a scene graph from images, several works that utilize axioms using fuzzy logic~\citep{hajek2013metamathematics} have been proposed because fuzzy logic can handle the intrinsic noise from the object detector. Specifically, there are algorithms for building SII graphs with a fuzzy logic ontology of spatial relations~\citep{atif2013explanatory, hudelot2008fuzzy}, an iterative message passing algorithm where the information about the objects maximizes the likelihood of the relationships~\citep{xu2017scene}, and the combination with Long Short-Term Memories~(LSTMs) for encoding the context given by the detected bounding boxes~\citep{zellers2018neural}. However, some of them are limited to only the spatial relationships and based on the implausible assumptions in real-work applications, for example, predicates must be mutually exclusive. 

There are several methods using a Conditional Random Field~(CRF) that encodes a fully connected graph and that labels or discards then nodes and edges by minimizing an energy function~\citep[e.g.,][]{kulkarni2013babytalk, chen2012understanding, chen2014detect}. However, some of them do not consider logical knowledge or use hand-crafted logical constraints, which are difficult to extend to other types of constraints.

Deep learning methods are also exploited for the task. Specifically, the detection of visual relationships using a Deep Relational Network~\citep{dai2017detecting}, a deep reinforcement learning model for detecting relationships and attributes~\citep{liang2017deep}, a message passing algorithm for sharing subject--object--predicate information among neural networks~\citep{li2017vip}, and an end-to-end system that can exploit the interaction of visual and geometric features of the subject, object and predicate~\citep{yin2018zoom}. However, the above systems cannot utilize the visual/geometric features of the subject/object and additional background knowledge together. 

A joint embedding with visual knowledge can exploit background knowledge. Implication, mutual exclusivity, and \emph{type-of} are exploited as logical constraints by \citet{ramanathan2015learning}, whereas a word embedding of the subject/object labels is used as background knowledge by \citet{lu2016visual}. \Citet{yu2017visual} use background knowledge in the form of a probability distribution of a relationship given the subject/object. However, the above methods do not exploit any type of logical constraints.

LTN can compensate for the shortcomings of all the studies mentioned above. LTNs allow multiple edges between nodes and can exploit the integration of the visual/geometric features of the subject/object with additional background knowledge encoded by logical constraints, which none of the above works can include. Therefore, we mainly compare the performance of RWFN to LTN, which we describe in detail next.

\subsection{Logic Tensor Networks}

Here, we describe basic concepts underlying Logic Tensor Networks~(LTNs)~\citep{serafini2016learning}. Although their structure is fundamentally different, RWFNs and LTNs use the same approach for mapping logical symbols to numerical values and learning reasoning relations among real-valued vectors using the logical formulas. However, LTNs combine reasoning with first-order logic with~(in contrast to WRFN) learning based on Neural Tensor Netowrks~(NTNs)~\citep{socher2013reasoning}. Consequently, the LTN framework can be implemented in TensorFlow~\citep{badreddine2021logic}.

We first define a First-Order-Logic~(FOL) language $\mathcal{L}$ and its
signature as containing three disjoint sets: i)~$\mathcal{C}$ (constants), ii)~$\mathcal{F}$ (functions), and iii)
$\mathcal{P}$ (predicate). However, we will not specify the function symbols $\mathcal{F}$ in detail because they are not used in the tasks. For specifying an \emph{arity} for a predicate symbol $s$, we use the notation as $\alpha(s)$. The logical formulas in $\mathcal{L}$ are used to describe relational knowledge. The objects in FOL are mapped to an interpretation domain $\subseteq \mathbb{R}^{n}$ so that every object is associated with an $n$-dimensional vector of real numbers and is reasoned over FOL. Intuitively, this $n$-tuple symbolizes $n$ numerical features of an object, and predicates are represented as fuzzy relationships on real vectors. With this numerical background, we can now establish the numerical \emph{grounding} of FOL with the following semantics. The term \emph{grounding} is used as a synonym of logical interpretation in a real world and needs to capture the latent correlation between the features of objects and their categorical or relational properties.

Let $n \in \mathbb{N}$. An \emph{$n$-grounding}, or simply \emph{grounding},
$\mathcal{G}$ for a FOL $\mathcal{L}$ is a function defined on the
signature of $\mathcal{L}$ satisfying the following conditions:
\begin{itemize}
    \item $\mathcal{G}(c) \in \mathbb{R}^n$ for every constant symbol $c
        \in \mathcal{C}$
    \item $\mathcal{G}(P) \in \mathbb{R}^{n \cdot \alpha(f)} \to [0,1]$
        for predicate sym.~$P \in \mathcal{P}$
\end{itemize}
Given a grounding $\mathcal{G}$, we can define the semantics of closed terms and
atomic formulas as follows:
\begin{gather*}
    \mathcal{G}(P(t_1, \dots, t_m)) \triangleq \mathcal{G}(P)(\mathcal{G}(t_1),
        \dots, \mathcal{G}(t_m))
\end{gather*}
The semantics for connectives, such as $\mathcal{G}(\neg \phi),
\mathcal{G}(\phi \land \psi), \mathcal{G}(\phi \lor \psi)$, and
$\mathcal{G}(\phi \rightarrow \psi)$, can be computed by following the
fuzzy logic such as the Lukasiewicz
$t$-norm~\citep{bergmann2008introduction}.

The grounding of an $m$-ary predicate $P$, namely
$\mathcal{G}(P)$, is defined as a generalization of the
NTN~\citep{socher2013reasoning}, as a function from $\mathbb{R}^{mn}$ to
$[0,1]$, as follows:
\begin{equation}
    \mathcal{G}_{\text{LTN}}(P)(\textbf{v})
    = \sigma (u_{P}^{\top} \texttt{f} (\textbf{v}^{\top} W_{P}^{[1:k]} \textbf{v} + V_{P} \textbf{v} + b_P))
    \label{eq:ltn_predicate}
\end{equation}
where $\textbf{v} = \langle \textbf{v}_1^\top, \dots, \textbf{v}_m^\top
\rangle^\top$ is the $mn$-ary vector obtained by concatenating each $\textbf{\texttt{v}}_i$. $\sigma$ is the sigmoidal logistic function, and $\texttt{f}$ is the hyperbolic tangent ($\tanh$). The parameters for $P$ are: $W_{P}^{[1:k]}$, a 3-D tensor in $\mathbb{R}^{k \times mn \times mn}, V_{P} \in \mathbb{R}^{k \times mn}, b_P \in \mathbb{R}^{k}$ and $u_P \in \mathbb{R}^k$. Because the RWFN model is a novel way of grounding a predicate as $\mathcal{G}_{\text{RWFN}}(P)$, we can directly compare the performance of RWFNs for the visual relationship detection tasks with LTNs.

The optimization of the truth values of the formulas in a LTN's knowledge base is directly involved with learning the groundings, i.e. \emph{grounded theory}. A \emph{partial grounding} $\hat{\mathcal{G}}$ is a grounding that can be defined on a subset of the signature of $\mathcal{L}$. A grounding $\mathcal{G}$ is said to be a completion of $\hat{\mathcal{G}}$ if $\mathcal{G}$ is a grounding for
$\mathcal{L}$ and coincides with $\hat{\mathcal{G}}$ on the symbols
where $\hat{\mathcal{G}}$ is defined. Let \emph{GT} be a grounded theory
which is a pair $\langle \mathcal{K}, \hat{\mathcal{G}}\rangle$ with a
set $\mathcal{K}$ of closed formulas and a partial grounding
$\hat{\mathcal{G}}$. A grounding $\mathcal{G}$ satisfies a \emph{GT}
$\langle \mathcal{K}, \hat{\mathcal{G}}\rangle$ if $\mathcal{G}$
completes $\hat{\mathcal{G}}$ and $\mathcal{G}(\phi)=1$ for all $\phi
\in \mathcal{K}$. A \emph{GT} $\langle \mathcal{K},
\hat{\mathcal{G}}\rangle$ is satisfiable if there exists a grounding
$\mathcal{G}$ that satisfies $\langle \mathcal{K},
\hat{\mathcal{G}}\rangle$. That is, deciding the satisfiability
of $\langle \mathcal{K}, \hat{\mathcal{G}}\rangle$ amounts to searching
for a grounding $\mathcal{G}$ such that all the formulas of
$\mathcal{K}$ are mapped to 1. If a \emph{GT} is not satisfiable, the
best possible satisfaction that we can reach with a grounding is of our
interest.

Therefore, the best-satisfiability problem is an optimization problem on the set of LTN parameters $\Theta \triangleq \{ W_P , V_P , b_P , u_P | P \in \mathcal{P} \}$ from Eq.~(\ref{eq:ltn_predicate}) to be learned. We use notation $\mathcal{G}(\cdot|\Theta)$ to indicate the grounding with grounding-function parameters equal to $\Theta$. Then, we define the best-satisfiability problem as finding the best parameter set:
\begin{equation}
    \Theta^{*} \triangleq \arg\max_{\Theta} \mathcal{G}(\land_{\phi \in \mathcal{K}} \phi|\Theta) - \lambda \lVert \Theta \rVert^{2}_{2},
    \label{eq:ltn_optimization}
\end{equation}
where $\lambda \lVert \Theta \rVert^{2}_{2}$ is a regularization term.

\section{Randomly Weighted Feature Networks}
The Randomly Weighted Feature Network~(RWFN)~\citep{hong2021insect} is a single-hidden-layer neural network that incorporates randomly drawn, untrained weights in an encoder that uses an adapted linear model as a decoder. 
A key characteristic of RWFN is the generation of unique hidden representations through the integration of two randomized methods for projecting from the input space to a higher-order space. 
Consequently, RWFNs can efficiently learn the degree of relationship among inputs by training only a linear decoder model. In addition, because all the weights of an encoder in the model are randomly drawn, the encoder can be shared with other classifiers that allow training only a decoder, which drastically reduces the space complexity of the model. 

\subsection{Unique Hidden Representation in RWFN}

\paragraph{Insect-brain--inspired Representation.}
In the insect brain, processed olfactory, visual, and mechanosensory stimuli are conveyed to the \emph{Mushroom Body~(MB)}~\citep{mobbs1982brain}, which can be viewed as the critical region responsible for multimodal associative learning~\citep{menzel2001searching}. In the fruit-fly brain, each of thousands of Kenyon Cells~(KCs) in the MB takes input from a random set of $\sim$7 inputs from Input Neurons~(INs)~\citep{caron2013random, inada2017origins}. Thus, a simplified neural circuit modeling the MB is a neural network with three layers consisting of: i)~INs that provide extracted features from olfactory, visual, and mechanosensory inputs, ii)~KCs generating the sparse-encoding of sensory stimuli, and iii)~mushroom body Extrinsic Neurons~(ENs) for activating several different behavioral responses~\citep{cope2018abstract}.
%
\Citet{hong2021insect} focused on the olfactory pathway between features extracted and coded by the AL and KCs in the MB~\citep{cope2018abstract, peng2017simple}; their RWFN architecture mimics the input transformation of odorant representation in the AL to the higher-order representation across the KCs in the MB.

\paragraph{Random Fourier Features.}
Kernel machines have attracted significant interest due to their capability of approximating functions with excellent performance for detecting decision boundaries given enough training data. These methods leverage transformations, via a lifting function $\phi$, that allows better discrimination among different inputs. Given dataset vector inputs $\textbf{x}, \textbf{y}
\in \mathbb{R}^d$, the kernel function $k(\textbf{x}, \textbf{y}) =
\langle\phi(\textbf{x}), \phi(\textbf{y})\rangle$ represents the
similarity (i.e., inner product) between $\textbf{x}$ and $\textbf{y}$
in the $\phi$-transformed space. However, because of the potential complexity
of the transformation $\phi$, learning the kernel function $k$ may suffer significant inefficiency of computational and storage costs.

To mitigate these kernel-machine costs, random Fourier features~\citep{rahimi2007random} provide a data transformation that permits a far less expensive approximation of the kernel function. For each vector input $\textbf{x} \in \mathbb{R}^d$, the method applies a randomized feature function $\textbf{z}: \mathbb{R}^d \to \mathbb{R}^D$ (generally, $D \gg d$ with
sample size $N \gg D$) that maps $\textbf{x}$ to evaluations of $D$
random Fourier bases from the Fourier transform of kernel $k$. In this
transformed space, linear operations can be used to approximate kernel evaluations, as in:
\begin{equation}
    k(\textbf{x}, \textbf{y})
        = \langle \phi(\textbf{x}), \phi(\textbf{y}) \rangle
        \approx \textbf{z}(\textbf{x})^{\top}\textbf{z}(\textbf{y})
    \label{eq:random_fourier_feature}
\end{equation}
Therefore, by transforming the input with $\textbf{z}$, fast linear learning can be used to approximate the evaluations of nonlinear kernel machines.
\Citet{hong2021insect} used random Fourier features as latent representations
that reduce the complexity of learning relations among real-valued entities. 

\subsection{Model Architecture}
\begin{figure}[t]\centering%
    \includegraphics[width=0.49\textwidth]{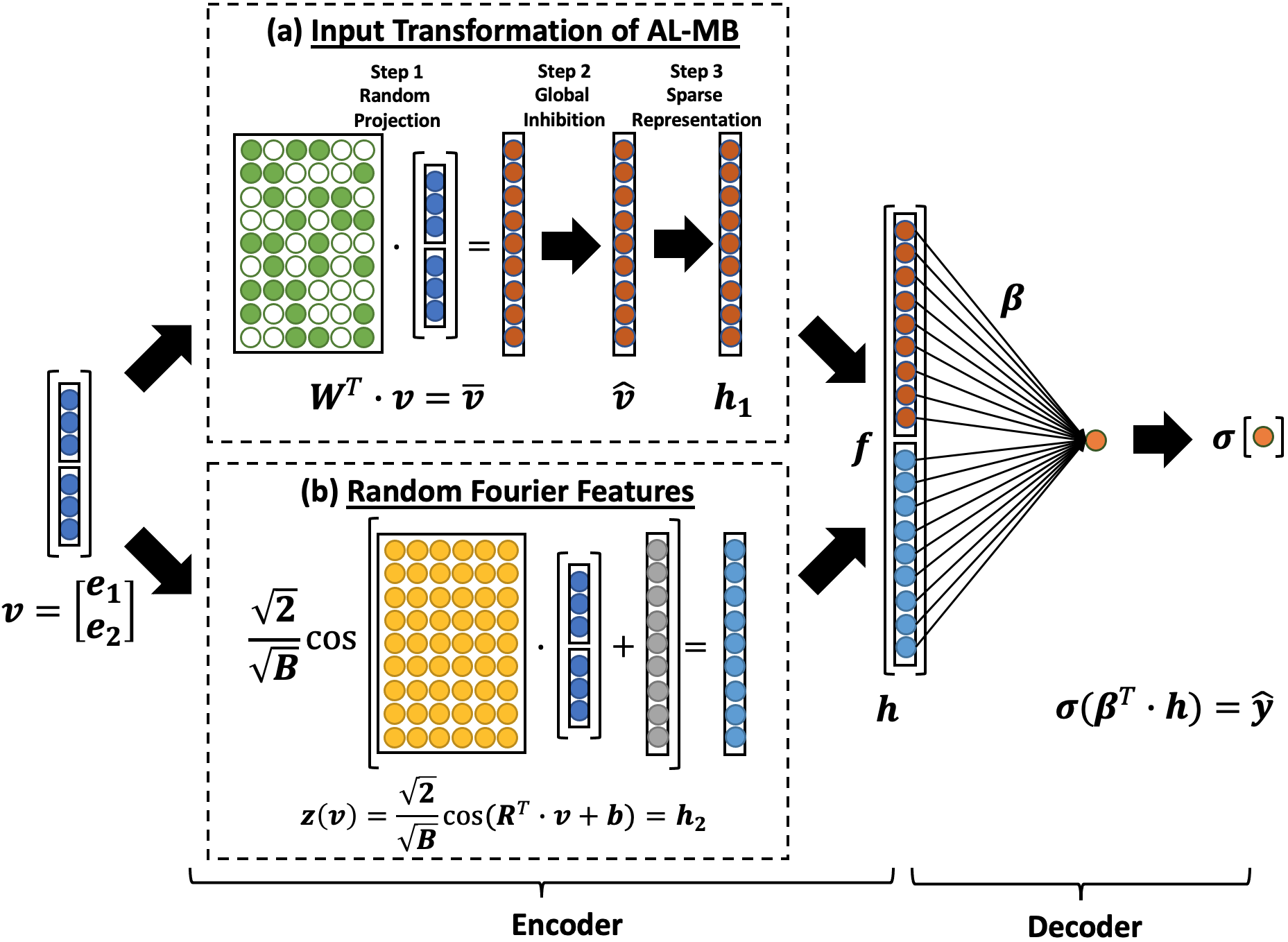}
    \caption{Visualization of the structure of the Randomly Weighted
    Feature Network. In the depicted case, the input vector $\textbf{v}$
    constitutes of two entities, $e_{1}, e_{2} \in \mathbb{R}^{3}$ and it
    shows to learn a binary relation between them $(e_{1}, R, e_{2})$, such
    as (Cat, hasPart, Tail)~\citep{hong2021insect}}
    \label{fig:rwfn-structure}
\end{figure}
\begin{figure}[t]\centering%
    \includegraphics[width=0.49\textwidth]{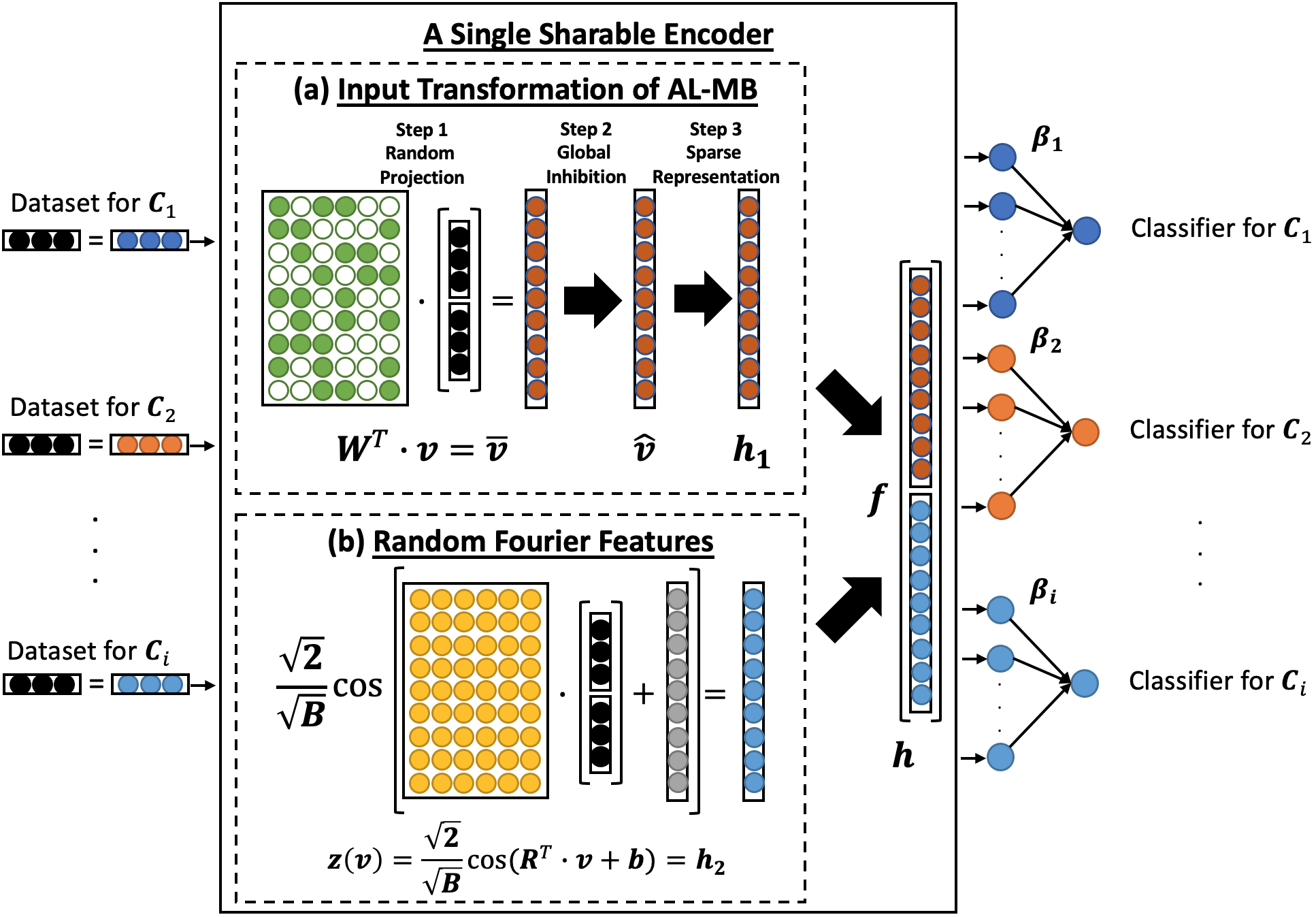}
    \caption{Visualization of the structure of the Randomly Weighted Feature
    Network with weight sharing. In the case of learning each classifier
    from the class $\mathcal{C}_{1}$ to the class $\mathcal{C}_{i}$, RWFNs
    allow us to use the same encoder to extract features from each data from the class $\mathcal{C}_{1}$ to the class $\mathcal{C}_{i}$~\citep{hong2021insect}}
    \label{fig:rwfn_ws_structure}
\end{figure}
Let the input vector $\textbf{v} = [\textbf{v}_1^{\top}, \dots,
\textbf{v}_m^{\top}]^{\top}$ be the $mn$-ary vector where $m$ is arity and $n$ is the input dimension, and consider a hidden node $j \in \{1,...,B\}$. To mimic the random AL--MB connections in the insect brain~\citep{hong2021insect}, hidden node~$j$ is associated with a random subset $\mathcal{S}_j \subset \{1,...,mn\}$ of $N_{in}=7$ of the $mn$ inputs (i.e., $|\mathcal{S}_j|=N_{in}=7$). Furthermore, hidden node~$j$ takes as input a binary-weighted combination of input nodes so that $w_{j,i}=1$ if $i \in S_j$ and $w_{j,i}=0$ otherwise. So, the input nodes are effectively gated by the weights on each hidden node, and the weight matrix $\textbf{W} \in \mathbb{R}^{mn \times B}$ in this computation is random, binary, and sparse. For the $j$th hidden node, $\Bar{v}_{j}$ represents the corresponding binary-weighted sum.

Next, for each hidden node $j$, a post-processing step produces the $j$th intermediate output $\hat{v}_{j}$ by subtracting the average of the binary-weighted sums, as in:
\begin{equation}
    \hat{v}_{j} = \Bar{v}_{j} - \frac{1}{B}\sum_{i=1}^{B} \Bar{v}_{i}
    \label{eq:kcnet_intermediate}
\end{equation}
where $B$ is the number of hidden units. Therefore, the sparse output of the $j$-th KC node can be defined as
    $h_{j}^{(1)} = g(\hat{v}_{j})$
where $g$ is the ReLU function~\citep{glorot2011deep} that allows the
model to produce sparse hidden output. By doing so, the output vector is defined as $\textbf{h}_{1} =
[h_{1}^{(1)}, \dots, h_{B}^{(1)}]^{\top}$. 

On the other hand, to generate random Fourier features, the randomized feature function $\textbf{z}(\cdot)$~\citep{rahimi2007random, sutherland2015error} was used so the inputs are projected as follows:
\begin{equation}
    \textbf{h}_{2} = \textbf{z}(\textbf{v}) = (\sqrt{2}/\sqrt{B}) \cos(\textbf{R}^{\top}\textbf{v} + \textbf{b})
\label{eq:rwfn_second_rep}
\end{equation}
where $\textbf{R} \sim \mathcal{N}^{mn \times B}(0,1)$ and $\textbf{b} \sim
U^{B}(0, 2\pi)$, which is Gaussian kernel approximation.
Consequently, the output vector $\textbf{h}_{2}$ can be considered as another latent representation of relationship among input.

Finally, using the above two latent representations, the RWFNs can be
defined as a function from $\mathbb{R}^{mn}$ to $[0,1]$:
\begin{equation}
    \mathcal{G}_{\text{RWFN}}(P)(\textbf{\texttt{v}})
    = \sigma \left(
        \boldsymbol\beta^\top
        \textbf{h}
    \right) \\
    = \sigma \left(
        \boldsymbol\beta^\top \texttt{f}\left(
            \begin{bmatrix}
                \textbf{h}_{1} \\
                \textbf{h}_{2}
            \end{bmatrix}
        \right)
    \right)
    \label{eq:rwfn}
\end{equation}
where $\textbf{h}$ is the final hidden representation obtained by applying (for numerical stabilization) the hyperbolic tangent~($\tanh$) function $\texttt{f}$ to the concatenation of $\textbf{h}_{1}$ and $\textbf{h}_{2}$, and $\sigma$ is the sigmoidal logistic function.
Because RWFN requires to adapt only $\boldsymbol\beta \in
\mathbb{R}^{2B}$, it possess a faster learning process with fewer
parameters compared to LTNs. Fig.~\ref{fig:rwfn-structure} shows a
visualization of the model structure.

\subsection{RWFNs with Weight Sharing}

Because of the above characteristics of the model, the randomized encoder of an RWFN can serve as a shared resource for multiple relatively simple (i.e., linear) downstream decoders trained for different classifiers. \citet{hong2021insect} called this property as \emph{weight
sharing}. Fig.~\ref{fig:rwfn_ws_structure} shows a visualization of the
structure of RWFN applied with weight sharing to the learning of
$i$ different classifiers. \Citeauthor{hong2021insect} show that RWFN with weight sharing allows RWFN to drastically reduce space complexity because the most costly components of the RWFN (i.e., the input and hidden layers) can be reused across multiple classifiers instead of being implemented (and trained) in parallel.

\section{RWFN for Visual Relationship Detection}
\Citet{donadello2019compensating} defined how to encode the problem of detecting visual relationship with LTNs. Because RWFNs can be readily applied using the same problem encodings of LTNs, we re-use the formalization of the problem from~\citeauthor{donadello2019compensating} for visual relationship detection in RWFNs. In this section, we briefly introduce that formalization.

\paragraph{The Knowledge base $\mathcal{K}_{\text{SII}}$.}
Given a dataset of images, let $B(p)$ be the corresponding set of bounding boxes of an image $p$ in the dataset. Each bounding box in $B(p)$ has its annotations about a set of labels that describe the contained physical object, and pairs of bounding boxes have their annotations about the semantic relations between the contained physical objects. Let $\Sigma_{\text{SII}} = \langle \mathcal{P}, \mathcal{C} \rangle$ be the signature where $\mathcal{P} = \mathcal{P}_{1} \cup \mathcal{P}_{2}$ is the set of predicates. $\mathcal{P}_{1}$ is a set of unary predicates indicating the \emph{object types} or \emph{semantic classes} for the label of bounding boxes. $\mathcal{P}_{2}$ includes binary predicates for the label of pairs of bounding boxes. Let $\mathcal{C} \triangleq \bigcup_{p} B(p)$ be the set of constants of all bounding boxes in the given dataset. A grounded theory can be defined as $\mathcal{T}_{\text{SII}} \triangleq \langle \mathcal{K}_{\text{SII}}, \hat{G}_{\text{SII}} \rangle$.

The knowledge base $\mathcal{K}_{\text{SII}}$ is for encoding the bounding box annotations in the dataset and some background knowledge about the domain so that it contains positive and negative examples (used for learning the grounding of the predicates in $\mathcal{P}$) and the background knowledge. The positive examples for a semantic class $C$ are the atomic formulas $C(b)$ for every bounding box $b$ labelled with class $C$. Regarding the negative examples, for a semantic class $C$, we consider the atomic formulas $\neg C(b)$ for every bounding box $b$ not labelled with $C$.
Regarding the background knowledge, \citet{donadello2019compensating} manually defined the logical constraints, such as \emph{negative domain and range constraints}, by referring to on-line linguistic resources such as FrameNet~\citep{baker1998berkeley} and VerbNet~\citep{schuler2005verbnet} that provide the range and domain of binary relations through the so-called frames data structure. 

\paragraph{Definition of New Features.}
On top of the features for grounding constants from \citet{donadello2017logic}, an extra set of features are added for representing constants in the knowledge base, called \emph{joint features}. Joint features include quantities such as: the inclusion ration of two bounding boxes, the area of intersection of two bounding boxes, the Euclidean distance between the centroids of two bounding boxes, and others that allow for better capturing the geometric interactions between two bounding boxes.

\paragraph{New Optimization for Knowledge Base.}
Equation~(\ref{eq:ltn_optimization}) defines how to learn the LTN parameters by maximizing the grounding of the conjunctions of the formulas in the knowledge base. However, there is a need to mitigate the issues that many $t$-norms have, such as leading the knowledge base satisfiability to zero, getting stuck in local optima, and underflow problems~\citep{donadello2019compensating}. Consequently, a mean operator was added to Eq.~(\ref{eq:ltn_optimization}) to return a global satisfiability of the knowledge base $\mathcal{K}_{\text{SII}}$ as follows:
\begin{equation}
    \Theta^{*} = \arg\max_{\Theta} \operatorname{mean}_{p} (\hat{\mathcal{G}}_{\text{SII}}(\phi | \Theta) | \phi \in \mathcal{K}_{\text{SII}}) - \lambda \lVert \Theta \rVert^{2}_{2}
    \label{eq:ltn_new_optimization}
\end{equation}
with $p \in \mathbb{Z}$.

\paragraph{Post Processing.} After a grounded theory $\mathcal{T}_{\text{SII}}$ is learned, the set of groundings $\hat{\mathcal{G}}_{\text{SII}}(r(b, b'))_{r \in \mathcal{P}_{2}}$ with a new pair of bounding boxes $\langle b, b' \rangle$ are computed. Then, every resulting grounding $\hat{\mathcal{G}}_{\text{SII}}(r(b, b'))$ is multiplied with the frequency of the predicate in the training set. Furthermore, equivalences between the binary predicates are exploited in order to normalize the groundings.

\section{Experiments}

In our experiments\footnote{All the source codes, models, and figures are available on \url{https://github.com/PavlicLab/AAAI2022-CLeaR2022-Visual_Relationship_Detection-RWFN}.}, we use the Visual Relationship Dataset~(VRD)~\citep{lu2016visual}, including 4000 images for the training set and 1000 for the testing set annotated with visual relationships. Bounding boxes are annotated with labels containing 100 unary predicates. These labels are for indicating animals, vehicles, clothes, and generic objects. Pairs of bounding boxes are annotated with labels including 70 binary predicates. These labels represent actions, prepositions, spatial relations, comparatives, or preposition phrases. The dataset contains 37993 instances of visual relationships and 6672 types of relationships. 1877 relationships occur only in the test set and are used to evaluate the zero-shot learning scenario.

\subsection{Methods}

\paragraph{VRD Tasks.}
The performances of both RWFNs and LTNs are tested on the following VRD standard tasks.
\begin{itemize}
    \item \emph{Phrase detection.} This task predicts a correct triple $\langle s, p, o \rangle$ and its location in a single bounding box that contains both the subject and the object. If the labels are the same as the ground truth triple and the predicted bounding box has at least 50\% overlap with a corresponding bounding box in the ground truth, the triple is a true positive. The ground truth bounding box means the union of the ground truth bounding boxes of the subject and the object. 
    \item \emph{Relationship detection.} This task predicts a correct triple/relationship and the bounding boxes that include the subject and the object of their relationship. If both bounding boxes overlap at least 50\% of the corresponding ones in the ground truth, the triple is a true positive. In addition, the labels for the predicted triple must match with the corresponding ones in the ground truth.
    \item \emph{Predicate detection.} For a given set of bounding boxes, this task predicts a set of correct binary predicates between them. Because the prediction does not depend on the performance of an object detector, the performances of LTNs/RWFNs are determined by their abilities to predict binary predicates, which is our interest.
\end{itemize}

\paragraph{Comparison of RWFNs and LTNs.}
The task is to complete the partial knowledge in the dataset by finding a grounding $\mathcal{G}^{*}_{\text{SII}}$. Therefore, for LTNs, it extends $\hat{\mathcal{G}}_{\text{LTN}}$ using Eq.~(\ref{eq:ltn_predicate}) such that:
\begin{align*}
    \mathcal{G}^{*}_{\text{SII}}(\mathcal{C}(b)) \to [0,1]
    \quad \text{and} \quad
    \mathcal{G}^{*}_{\text{SII}}(\mathcal{R}(b_{1}, b_{2})) \to [0,1]
\end{align*}
for every unary predicate $\mathcal{C}$ and binary predicate $\mathcal{R}$ and every (pair of) bounding box in the dataset. Because for RWFNs $\hat{\mathcal{G}}_{\text{RWFN}}$ using Eq.~(\ref{eq:rwfn}) is instead used for grounding unary and binary predicates, we can directly compare the performances between LTNs and RWFNs.

\Citet{donadello2019compensating} test the performance of LTNs with two grounded theories, $\mathcal{LTN}^{\text{expl}}$ and $\mathcal{LTN}^{\text{prior}}$. The first one indicates $\mathcal{LTN}^{\text{expl}} = \langle \mathcal{K}_{\text{expl}}, \hat{\mathcal{G}}_{\text{LTN}} \rangle$ where $\mathcal{K}_{\text{expl}}$ includes only positive and negative examples for predicates predictions. The second grounded theory means $\mathcal{LTN}^{\text{prior}} = \langle \mathcal{K}_{\text{prior}}, \hat{\mathcal{G}}_{\text{LTN}} \rangle$ where $\mathcal{K}_{\text{prior}}$ contains examples as well as logical constraints. Therefore, we can check the contribution of using logical constraints. By referring to the above setting of LTNs, the performance of RWFNs can also be evaluated by creating two grounded theories, $\mathcal{RWFN}^{\text{expl}}$ and $\mathcal{RWFN}^{\text{prior}}$. Furthermore, we tested RWFNs with weight sharing by building two grounded theories, $\mathcal{RWFN_{WS}}^{\text{expl}}$ and $\mathcal{RWFN_{WS}}^{\text{prior}}$. In the setting for RWFNs, all predicates have their own encoders whereas in the setting for RWFNs with weight sharing, we created one predefined encoder for unary predicate and another for binary predicate in advance, and those encoders were used as the shared encoders for grounding predicates. Thus, we can readily confirm how the shareable encoders contribute to achieve the performance. 

We first train all the models~-- RWFNs, RWFNs with weight sharing, and LTNs~-- on the VRD training set, and then we test them on the VRD test set. All the models have been evaluated their abilities to generalize to the 1877 relationships never seen in the training phase.

\paragraph{Evaluation Metric.}
Following \citet{donadello2019compensating}, for each image in the test set, we use the grounded theories for each model to compute the ranked set of groundings with a pair of bounding boxes computed with an object detector (the R-CNN model from~\citet{lu2016visual}) or taken from the ground truth (for the predicate detection). We then use the recall@100/50~\citep{lu2016visual} as evaluation metrics because the annotation is not complete, and precision would wrongly penalize true positives. In addition, every pair of bounding boxes with all the binary predicates are classified as many predicates can occur between two objects, and it is not always possible to define a preference between predicates. This choice is counterbalanced by predicting the correct relationships within the top 100 and 50 positions.

\paragraph{Hyperparameter Setting.}
For hyperparameter setting of LTNs, we followed the setting of~\citet{donadello2019compensating}.
In equation~(\ref{eq:ltn_new_optimization}), $p$ was set to $-1$~(harmonic mean). The chosen $t$-norm is the Lukasiewicz one. The number of tensor layers in Eq.~(\ref{eq:ltn_predicate}) is set as $k = 5$ and $\lambda = 10^{-10}$ in Eq.~(\ref{eq:ltn_new_optimization}). In order to mimic the results from \citeauthor{donadello2019compensating}, the optimization is performed separately on $\mathcal{LTN}^{\text{expl}}$ and $\mathcal{LTN}^{\text{prior}}$ with 10000 training epochs of the RMSProp optimizer~\citep{Tieleman2012} in TensorFlow~\citep{abadi2016tensorflow}.

For hyperparameter setting of RWFNs and RWFNs with weight sharing, the value of $p$ for harmonic mean, $t$-norm, $\lambda$ and the number of training epochs are set as above. In Eq.~(\ref{eq:rwfn}), the size of $B$ is set to 500 for unary predicate whereas it is set to 1000 for binary predicate. In contrast with \citet{hong2021insect}, in order to train RWFNs and RWFNs with weight sharing, we use the Follow-The-Regularized-Leader~(FTRL) optimizer~\citep{mcmahan2013ad} instead of the RMSProp optimizer. It has been known that the FTRL optimizer is suitable for shallow models with large and sparse feature spaces, and this is properly applicable to optimize RWFNs because the feature representations in RWFNs can be sparse due to the insect-brain--inspired representation. For the hyperparameters of the FTRL optimizer, the learning rate is set to 1, and $g, \lambda_1$ and $\lambda_2$ are set to the standard values that TensorFlow provides, where are $-0.5, 0,$ and $0$, respectively. See the Appendix for additional experimental details.

\subsection{Results}
Table~\ref{tab:results} shows the results of comparison between LTNs, RWFNs, and RWFNs with weight sharing.
\begin{table*}[t!]
    \centering
    \begin{tabular}{ccccccc}
    \toprule
    Task: & Phrase Det. & Phrase Det. & Relation Det. & Relation Det. & Predicate Det. & Predicate Det. \\
    \cmidrule(lr){1-1}\cmidrule(lr){2-3}\cmidrule(lr){4-5}\cmidrule(lr){6-7}
    Evaluation: & R@100 & R@50 & R@100 & R@50 & R@100 & R@50 \\
    \cmidrule(lr){1-1}\cmidrule(lr){2-3}\cmidrule(lr){4-5}\cmidrule(lr){6-7}
    \morecmidrules
    \cmidrule(lr){1-1}\cmidrule(lr){2-3}\cmidrule(lr){4-5}\cmidrule(lr){6-7}
    $\mathcal{LTN}^{\text{expl}}$ & \emph{14.99 $\pm$ 1.2} & 10.52 $\pm$ 0.59 & \emph{13.42 $\pm$ 1.1} & 9.39 $\pm$ 0.39 & 68.59 $\pm$ 0.93 & 50.61 $\pm$ 1.7 \\
    $\mathcal{LTN}^{\text{prior}}$ & \textbf{16.15 $\pm$ 0.75} & \textbf{11.82 $\pm$ 0.37} & \textbf{14.81 $\pm$ 0.81} & \textbf{10.69 $\pm$ 0.32} & 74.66 $\pm$ 1.6 & \emph{54.9 $\pm$ 2.1}\\
    \cmidrule(lr){1-1}\cmidrule(lr){2-3}\cmidrule(lr){4-5}\cmidrule(lr){6-7}
    $\mathcal{RWFN}^{\text{expl}}$ & \emph{15.48 $\pm$ 0.51} & 10.7 $\pm$ 0.74 & \emph{14.13 $\pm$ 0.51} & 9.61 $\pm$ 0.61 & \emph{76.97 $\pm$ 0.08} & \emph{56.17 $\pm$ 0.61} \\
    $\mathcal{RWFN}^{\text{prior}}$ & \emph{15.93 $\pm$ 0.39} & \emph{10.78 $\pm$ 0.81} & \emph{14.44 $\pm$ 0.47} & \emph{9.89 $\pm$ 0.65} & \emph{77.3 $\pm$ 0.6} & \textbf{57.01 $\pm$ 1.3} \\
    $\mathcal{RWFN_{WS}}^{\text{expl}}$ & \emph{15.19 $\pm$ 0.31} & 10.45 $\pm$ 0.84 & \emph{14.03 $\pm$ 0.36} & \emph{9.49 $\pm$ 0.99} & \emph{77.11 $\pm$ 1.4} & \emph{55.62 $\pm$ 0.91} \\
    $\mathcal{RWFN_{WS}}^{\text{prior}}$ & \emph{15.67 $\pm$ 0.75} & \emph{11.1 $\pm$ 0.99} & \emph{14.4 $\pm$ 0.49} & \emph{10.08 $\pm$ 0.85} & \textbf{77.33 $\pm$ 0.32} & \emph{56.89 $\pm$ 0.91} \\
    \bottomrule
    \end{tabular}
    \caption{Results on Visual Relationship Dataset (R@N: Recall at N) for LTN, RWFN, and RWFN with weight sharing. 95\% CIs (i.e., $\text{MEAN} \pm 2 \times \text{SD}$) shown for all models. Best-mean CIs shown in \textbf{bold}; CIs that overlap with best-mean CI in \emph{italic.}}
    \label{tab:results}
\end{table*}
We performed the training 5 times obtaining 5 models for $(\cdot)^{\text{expl}}$ and $(\cdot)^{\text{prior}}$, respectively. For each task and each grounded theory, we report 95\% confidence intervals~(CIs) on the means for the sample results given by these models.

Most differences in performance between RWFN variants (with and without weight sharing) and LTNs are marginal. Across all tasks, whenever LTNs were shown to have higher sample mean performance, LTN CIs overlapped with CIs of both RWFNs variants, indicating no clear significant difference in performance. For predicate detection, RWFNs variants did show significantly better performance than LTNs. In particular, without the use of logical constraints, RWFNs and RWFNs with weight sharing achieved better performance for detecting predicates than LTNs with logical constraints. This indicates the randomized projections of inputs in RWFNs effectively capture and learn the visual relationships among inputs. Furthermore, RWFNs with weight sharing do not suffer a deterioriation in performance. In fact, making use logical constraints in RWFNs with weight sharing may be more beneficial than in RWFNs because the gap of the performance of the model with and without logical constraints is larger than the original RWFNs.  

\paragraph{Relative Complexity of RWFNs and LTNs.}
We compare the number of parameters for grounding a unary predicate for each model to comprehend the relative performance of RWFNs and LTNs.  The input dimension in the dataset for both RWFNs and
LTNs is $n = 105$ for grounding a unary predicate.
As shown in Eq.~(\ref{eq:ltn_predicate}), the
parameters to learn in LTNs are $\{ u_{P} \in \mathbb{R}^k,
W_{P}^{[1:k]} \in \mathbb{R}^{n \times n \times k}, V_{P} \in
\mathbb{R}^{k \times n}, b_P \in \mathbb{R}^k \}$, where $k = 5$
in the setting of the LTNs. Therefore, the number of parameters
in LTNs is $(n^2 + n + 2) \cdot k = (105^2 + 105 + 2) \cdot 5 = 55660$.
On the other hand, in Eq.~(\ref{eq:kcnet_intermediate}) and Eq.~(\ref{eq:rwfn_second_rep}), the number of parameters in
RWFNs are $\{ \textbf{W} \in \mathbb{R}^{n \times B}, \textbf{R} \in
\mathbb{R}^{n \times B}, \textbf{b} \in \mathbb{R}^{B}, \boldsymbol\beta
\in \mathbb{R}^{2B} \}$, where $B = 500$ in the setting of
the RWFNs. Thus, the number of parameters in RWFNs is $(2n + 3)
\cdot B = (2 \cdot 105 + 3) \cdot 500 = 106500$.
Although RWFNs require more space complexity compared to
LTNs, the parameters $\{ \textbf{W}, \textbf{R},
\textbf{b} \}$ in RWFNs are randomly drawn weights. Thus, it is necessary to compare the number of learnable parameters across the two models as well.

All of the above parameters of LTNs have to be learnable whereas the
parameters to learn in RWFNs for object type classification are only
$\boldsymbol\beta \in \mathbb{R}^{2B}$. Thus, the number of learnable
parameters is $1000$, which is much smaller than that of LTNs. It means
that the ratio of the two numbers of parameters to learn is about
$1000:55660 \approx 1:56$. Consequently, non-learnable parameters in
RWFNs can have significant potential to represent the latent relationship
among objects and efficiently extract relational
knowledge even though using fewer adaptable parameters. Furthermore, the number of LTN parameters heavily depends on the number of features, whereas RWFNs are independent of features. 

\paragraph{Space Complexity of RWFNs with Weight Sharing.}
The unique property of RWFNs, weight sharing, allows reducing space complexity greatly when multiple classifiers are used
simultaneously. By referring to the depicted case of learning $i$ classifiers in Fig.~\ref{fig:rwfn_ws_structure}, we compute the space complexity for RWFNs with and without weight sharing for the detection of unary predicates in the tasks. \Citet{hong2021insect} show that the space complexity for the original RWFNs is $(2 \cdot n + 3)\cdot B \cdot i = (2 \cdot 105 + 3) \cdot 500 \cdot 100 = 1.065 \times 10^7$ because the number of classifiers $i$ is 100. However, with weight sharing, RWFNs can achieve much better space complexity, which is $2 \cdot n \cdot B + B + 2 \cdot B \cdot i = 2 \cdot 105 \cdot 500 + 500 + 2 \cdot 500 \cdot 100 = 205500$ and this complexity is much smaller than one of the original RWFNs. Furthermore, because the LTNs require $55660$ parameters for grounding a single unary predicate, the space complexity of LTNs for grounding all unary predicates is $55660 \times 100 = 5566000$, which is much larger than the space complexity of RWFNs with weight sharing. It indicates that the ratio of the two space complexities between RWFNs with weight sharing and LTNs is $205500:5566000 \approx 1:27$ and that the weight sharing property allows RWFNs to be more cost efficient and economical than LTNs even though the performance of RWFNs with weight sharing for predicate detection is better than LTNs.

\section{Conclusion}
We showed that Randomly Weighted Feature Networks can be extended to a zero-shot approach that learns the similarity with other seen triples in the presence of logical background knowledge. The results on the Visual Relationship Dataset show that RWFNs outperform LTNs with far fewer parameters to train. The proposed method addresses not only an emerging problem in AI datasets due to the high annotation effort and their consequent incompleteness but also a critical problem in the neuro-symbolic domain~-- the reduction of the number of training parameters~-- which could allow for online training of the neuro-symbolic models with real-time performance possible in the future. In addition, the combination of bio-inspired neural models with logical prior knowledge shows how biologically inspired neural networks \emph{plus knowledge} can learn with few parameters with respect to artificial neural networks and can be applied to even more complex computer vision tasks. 

The proposed method can be improved in various ways. For one, RWFNs can be employed in tasks that should extract structural knowledge
from images as well as text, such as visual question answering
using Visual Genome dataset~\citep{krishna2016visual}. Moreover, other perspectives from neuroscience may lead to biologically plausible learning algorithms that might apply to further optimizations of RWFNs~\citep{krotov2019unsupervised, kasai2021spine, kappel2018dynamic}. Furthermore, RWFNs may be able to incorporate a recurrent component for representing dynamic features of time-series data, similar to reservoir computing~\citep{ferreira2009genetic, sun2017deep, wang2019echo}; this approach may allow for extracting time-varying relational knowledge necessary for developing a framework for data-driven reasoning over temporal logic.

\appendix
\section{Appendix: Details of Experiments}
\paragraph{Hyperparameter Searching for RWFNs}
We used the Optuna framework~\citep{akiba2019optuna} with 500 iterations in the range of $[64, 1024]$ to determine the best number of hidden nodes $\beta$ in Eq.~(\ref{eq:rwfn}). Because in the Optuna framework, we can formalize hyperparameter optimization as the maximization or minimization process of an objective function that takes a set of hyperparameters as input and returns a validation score, we can easily construct the parameter search space dynamically. In addition, the framework provides efficient sampling methods, such as relational sampling that exploits the correlations among the parameters.

\paragraph{Hardware specification of the server.}
The hardware specification of the server that we used to experiment is as follows:
\begin{itemize}
    \item CPU: Intel\textregistered{} Core\textsuperscript{TM} i7-6950X CPU @ 3.00GHz (up to 3.50 GHz)
    \item RAM: 128 GB (DDR4 2400MHz)
    \item GPU: NVIDIA GeForce Titan Xp GP102 (Pascal architecture, 3840 CUDA Cores @ 1.6 GHz, 384 bit bus width, 12 GB GDDR G5X memory)
\end{itemize}

\bibliography{aaai22}

\end{document}